\documentclass[preprint,12pt,authoryear]{elsarticle}
\usepackage{amssymb}
\usepackage{amsmath}
\usepackage{float}
\usepackage{booktabs}

\journal{Nuclear Physics B}

\begin{document}

\begin{frontmatter}

\title{ICON: Invariant Counterfactual Optimization with Neuro-Symbolic Priors for Text-Based Person Search}

% \author{}
% \affiliation{organization={},
%             addressline={}, 
%             city={},
%             postcode={}, 
%             state={},
%             country={} }

\author{Xiangyu Wang$^{1}$, Zhixin Lv$^{1}$, Yongjiao Sun$^{1,2}$, Anrui Han$^{1}$, Ye Yuan$^{3}$, Hangxu Ji$^{1,4}$}

% Affiliation 1
\affiliation{organization={1.College of Computer Science and Engineering, Northeastern University},
            % addressline={},
            city={Shenyang},
            postcode={110819},
            % state={},
            country={China}}

% Affiliation 2
\affiliation{organization={2.Key Laboratory of Intelligent Computing in Medical Image of Ministry of Education, Northeastern University},
            % addressline={},
            city={Shenyang},
            postcode={110819},
            % state={},
            country={China}}

% Affiliation 3
\affiliation{organization={3.School of AI, Beijing Institute of Technology},
            % addressline={},
            city={Beijing},
            postcode={100081},
            % state={},
            country={China}}

% Affiliation 4
\affiliation{organization={4.Foshan Graduate School of Innovation, Northeastern University},
            % addressline={},
            city={Foshan},
            postcode={528311},
            % state={},
            country={China}}

% Optional: Emails corresponding to the authors
% 2410833@stu.neu.edu.cn (Xiangyu Wang)
% lvzx1@mails.neu.edu.cn (Zhixin Lv)
% sunyongjiao@mail.neu.edu.cn (Yongjiao Sun)
% 13504263512@163.com (Anrui Han)
% yuan-ye@bit.edu.cn (Ye Yuan)
% jihangxu@mail.neu.edu.cn (Hangxu Ji)

\begin{abstract}
Text-Based Person Search (TBPS) holds unique value in real-world surveillance bridging visual perception and language understanding, yet current paradigms utilizing pre-training models often fail to transfer effectively to complex open-world scenarios. The reliance on "Passive Observation" leads to multifaceted spurious correlations and spatial semantic misalignment, causing a lack of robustness against distribution shifts. To fundamentally resolve these defects, this paper proposes ICON (Invariant Counterfactual Optimization with Neuro-symbolic priors), a framework integrating causal and topological priors. First, we introduce Rule-Guided Spatial Intervention to strictly penalize sensitivity to bounding box noise, forcibly severing location shortcuts to achieve geometric invariance. Second, Counterfactual Context Disentanglement is implemented via semantic-driven background transplantation, compelling the model to ignore background interference for environmental independence. Then, we employ Saliency-Driven Semantic Regularization with adaptive masking to resolve local saliency bias and guarantee holistic completeness. Finally, Neuro-Symbolic Topological Alignment utilizes neuro-symbolic priors to constrain feature matching, ensuring activated regions are topologically consistent with human structural logic. Experimental results demonstrate that ICON not only maintains leading performance on standard benchmarks but also exhibits exceptional robustness against occlusion, background interference, and localization noise. This approach effectively advances the field by shifting from fitting statistical co-occurrences to learning causal invariance.
\end{abstract}


\end{frontmatter}

\section{Introduction}

Text-Based Person Search (TBPS) serves as a pivotal foundational task bridging visual perception and language understanding. Its objective is to retrieve and localize specific targets from untrimmed gallery images based on natural language descriptions. Unlike Image-Based Person Search (IBPS), which relies on visual priors of the target, TBPS is not constrained by the absence of visual query samples. Consequently, it holds unique application value in real-world scenarios, such as surveillance retrieval based on verbal descriptions and interactive security systems. With the advent of large-scale vision-language pre-training models, this field has witnessed significant progress. Mainstream approaches typically adopt a dual-tower encoder architecture, utilizing pre-trained models like CLIP\citep{radford2021clip} as the backbone to map visual and textual modalities into a unified feature space via contrastive learning losses. However, while existing models achieve exceptional performance on closed-set benchmarks such as PRW and CUHK-SYSU, this superiority often fails to transfer effectively to complex open-world scenarios. The stark performance disparity between metric saturation in closed testing environments and the lack of robustness when facing distribution shifts in practical applications exposes the inherent defects of current paradigms. Therefore, moving beyond mere metric improvements to construct robust retrieval models capable of maintaining semantic consistency under complex environments and distribution shifts has become a critical scientific problem urgently awaiting resolution in this field.

\begin{figure}[H] %H为当前位置，!htb为忽略美学标准，htbp为浮动图形
		\centering %图片居中
		\includegraphics[width=0.8\textwidth]{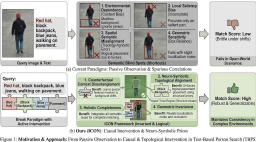} 
		\caption{This is Intro} %最终文档中希望显示的图片标题
		\label{Fig.overview} %用于文内引用的标签
	\end{figure}%结束环境

To enhance retrieval performance, existing research has conducted extensive exploration. Traditional TBPS methods primarily focus on extracting global and local features from images and text, achieving modality alignment through simple feature concatenation or weighting. While these methods laid the foundation for the field, they often struggle when processing fine-grained semantics and complex structured descriptions. To overcome the limitations of coarse-grained matching, recent research has gradually shifted towards deep cross-modal alignment. First, to capture alignment at a deeper level, SPT \citep{tan2024spt} and MGCC \citep{wu2024mgcc} introduced saliency guidance and multi-granularity contrastive learning, respectively, attempting to mine latent associations at the sub-pixel level. Second, addressing structured semantic challenges in complex queries, IRRA \citep{jiang2023irra} and MACA \citep{su2024maca} enhanced the model's discriminative ability regarding logical attributes through implicit relational reasoning. Furthermore, addressing the distribution discrepancy between training and inference data, ViPer \citep{zhang2025viper} innovatively introduced a multi-granularity visual perturbation mechanism to enhance model adaptability.

Nevertheless, although the aforementioned methods have significantly pushed the performance ceiling, they generally adhere to a \textit{Passive Observation} logic in their training paradigms—primarily relying on fitting high-frequency co-occurrence patterns within the data distribution. This over-reliance on superficial statistical signals leads to two dimensional Semantic Blind Spots: \textit{Multifaceted Spurious Correlations} and \textit{Spatial Semantic Misalignment}. \textbf{Blind Spot I: Multifaceted Spurious Correlations.} Due to a lack of active intervention, models are highly prone to the Clever Hans effect, which differentiates into three statistical shortcuts: (a) \textit{Geometric Sensitivity}, where the model relies excessively on precise bounding box positions; (b) \textit{Environmental Dependency}, where inference is based on background context (e.g., pavement) rather than pedestrian features; and (c) \textit{Local Saliency Bias}, where the model focuses solely on salient features (e.g., red clothing) while ignoring holistic completeness. \textbf{Blind Spot II: Spatial Semantic Misalignment.} Existing models are essentially Topology-Agnostic, ignoring the basic spatial inductive bias of the human body. This leads to logical errors driven by textural similarity (e.g., misidentifying black trousers on the legs as a black backpack). In summary, existing methods fail to fundamentally resolve the dependence on these statistical shortcuts from the perspective of fine-grained causal intervention and topological structure.

To break the \textit{Passive Observation} paradigm and integrate causal and topological priors into model design, this paper proposes \textbf{ICON} (Invariant Counterfactual Optimization with Neuro-symbolic priors). Our core hypothesis is that true robustness stems from the deep modeling of \textit{Semantic Causality} and \textit{Spatial Logic}. Therefore, ICON is built upon four fundamental principles: Geometric Invariance, Environmental Independence, Holistic Completeness, and Spatial Layout Consistency. Specifically, we construct a unified framework of complementary mechanisms to precisely target the identified blind spots. To achieve geometric invariance, we introduce a \textbf{Rule-Guided Spatial Intervention} that strictly penalizes sensitivity to bounding box noise, forcibly severing the Location Shortcut. Simultaneously, to ensure environmental independence, we implement \textbf{Counterfactual Context Disentanglement} by generating counterfactual samples via semantic-driven background transplantation, thereby compelling the model to ignore background interference. Furthermore, addressing local saliency bias, we employ \textbf{Saliency-Driven Semantic Regularization} with an adaptive masking strategy to compel the model to reconstruct semantics from remaining cues, guaranteeing holistic completeness. Finally, to resolve spatial semantic misalignment, we incorporate \textbf{Neuro-Symbolic Topological Alignment}, which utilizes neuro-symbolic priors to constrain the feature matching process, ensuring that activated regions are not only texturally similar but also topologically consistent with human structural logic.

The main contributions of this paper are summarized as follows: \begin{itemize} \item We systematically deconstruct the semantic blind spot problem in TBPS from the dual perspectives of causal intervention and topological logic. We identify, for the first time, three dimensions of spurious correlations and a fundamental lack of spatial logic as the core failure modes of the Passive Observation paradigm.

\item We propose the ICON framework, a novel approach based on neuro-symbolic counterfactual intervention. By integrating a 3+1 mechanism comprising three layers of counterfactual generation and a core neuro-symbolic topological alignment module, ICON effectively transforms the learning process from fitting statistical co-occurrences to learning causal invariance.

\item We conduct extensive comparative experiments on standard benchmarks and various designed distribution shift scenarios. Results demonstrate that ICON not only maintains leading performance on conventional metrics but also exhibits exceptional robustness and generalization capabilities against occlusion, background interference, and localization noise.

\end{itemize}

\section{Related Work}
\subsection{Occlusion in Scene-level Retrieval}
Occlusion is a central challenge for person retrieval in real-world scenes. It disrupts the integrity of a person’s appearance and causes global representations to pick up noise from occluders or background, which severely degrades matching performance. Early strategies primarily relied on data augmentation. Random Erasing, RE \citep{zhong2020re}, randomly selects a rectangular region during training and fills its pixels with random values. RE introduces no extra model parameters and requires no external annotations, yet it discourages over-reliance on a single salient cue and promotes learning multiple complementary, unoccluded local cues, leading to markedly improved robustness to occlusion. A second line of research designs explicit part-based architectures. PCB, Part-based Convolutional Baseline \citep{sun2018pcb}, uniformly partitions the backbone feature map along the horizontal axis and learns per-part embeddings, but this blind partition lacks semantic alignment. To obtain more precise body parts, subsequent work introduces pose estimation as prior knowledge. PGFA, Pose-Guided Feature Alignment \citep{miao2019pgfa}, uses a pretrained pose estimator to extract keypoints and derive visibility attention maps that guide the model to focus on visible body regions while filtering noise from occluders and background. Although effective, this approach depends on an external model whose reliability may drop under severe occlusion, risking cascading failures.

Recent advances move toward more realistic simulation, modern architectures, and multimodal integration. Researchers have noted a domain gap between RE’s gray blocks and real occluders. SPT, Saliency-Guided Patch Transfer \citep{tan2024spt}, synthesizes photorealistic occlusion by separating foreground persons from background obstacles via saliency detection and pasting the person onto scenes with real obstacles. In parallel, transformer-based training with masked image modeling encourages models not only to match the visible regions but also to reconstruct the occluded parts, yielding descriptors that degrade gracefully. Moreover, occlusion has extended to multimodal retrieval. MGCC \citep{wu2024mgcc} targets text-based re-identification under occlusion through fine-grained token selection and multi-granularity contrast. Pushing this challenge to scene-level retrieval, ViPer \citep{zhang2025viper} highlights the train–test mismatch: training typically observes clean, aligned crops, while inference encounters unavoidable visual perturbations including occlusion, background clutter, and boundary misalignment. This reframes occlusion as a core component of a broader perturbation problem that must be addressed systematically within an end-to-end multimodal pipeline.

\subsection{Person Search}

Person search aims to unify a more practical setting: given a query image of a target pedestrian, the system must detect and re-identify that person within an uncropped gallery of scene images. Early solutions followed a two-step paradigm that first runs a standalone pedestrian detector and then applies a separate re-identification model on detected proposals \citep{zheng2017prw}. In the same year, the OIM end-to-end framework was introduced \citep{xiao2017oim}. It jointly optimizes detection and re-identification within a single network and, together with the CUHK-SYSU dataset and the OIM loss, established a strong baseline for person search \citep{xiao2017oim}. The OIM loss maintains a dynamic look-up table and a circular queue to cope with the large number of identity classes, alleviating training difficulties of the softmax classifier. While the end-to-end paradigm improves efficiency, it exposes a fundamental objective conflict: detection seeks intra-class compactness to separate “person” from background, whereas re-identification seeks inter-class dispersion to distinguish identities; optimizing both on a shared backbone can yield suboptimal solutions.

To resolve this, NAE proposes a feature-level decoupling that decomposes embeddings into norm and angle: the norm is used as a detection confidence for person-versus-background, while the angular component (cosine similarity) handles identity discrimination, effectively orthogonalizing the two objectives in different embedding dimensions \citep{chen2020nae}. DMRNet further performs architectural decoupling with deep multi-task branches and a memory-reinforced mechanism to stabilize re-identification feature learning \citep{han2021dmrnet}. In terms of detection paradigms, OIM and NAE are built on two-stage, anchor-based Faster R-CNN-style designs \citep{xiao2017oim,chen2020nae}. With the rise of efficient one-stage detectors, AlignPS becomes the first single-stage, anchor-free person search framework and addresses feature–region misalignment induced by the anchor-free design \citep{yan2021alignps}. Following the transformer revolution in detection, PSTR formulates person search as sequence prediction and uses an encoder–decoder architecture to jointly output detection boxes and re-identification descriptors without NMS \citep{cao2022pstr}. The current frontier shifts to a more challenging multimodal task, text-based person search (TBPS), which retrieves the target using natural language descriptions. SDPG defines the task in full images and introduces PRW-TBPS and CUHK-SYSU-TBPS benchmarks while leveraging semantics to guide proposal generation \citep{zhang2024tbpsfull}. Building on this, MACA employs a coarse-to-fine alignment strategy with attribute-level cues to improve multimodal matching \citep{su2024maca}, and a carefully tuned CLIP baseline has been validated as a strong TBPS baseline \citep{cao2024tbpsclip}. Most recently, ViPer employs a visual-perturbation training scheme that explicitly simulates occlusion, boundary shifts, and background clutter during training to shrink the train–test gap, representing the latest advance in robust TBPS \citep{zhang2025viper}.

\section{Problem Definition and Motivation}

This section aims to provide a formal mathematical description of the robust Text-Based Person Search (TBPS) task we study.

First, we define the standard TBPS task. Given an image gallery with $N$ uncropped scene images and a set of natural-language queries:
\begin{equation}
\mathcal{G}=\{I_1,I_2,\ldots,I_N\},\qquad
\mathcal{Q}=\{q_1,q_2,\ldots,q_M\},
\end{equation}
where each query $q_j$ describes a target person.

For any given query $q\in\mathcal{Q}$, the goal of TBPS is to retrieve from the entire gallery $\mathcal{G}$ a ranked list of bounding boxes:
\begin{equation}
\mathcal{B}_{\mathrm{ranked}}(q)=[b_1,b_2,\ldots,b_K],\quad
b_k=(I_k,\mathbf{bb}_k),\ I_k\in\mathcal{G},\ \mathbf{bb}_k\in\mathbb{R}^4.
\end{equation}

The ground truth for $q$ is a set of positive bounding boxes:
\begin{equation}
\mathcal{B}_{gt}(q)\subset \bigcup_{i=1}^{N}\mathcal{B}_i,
\qquad
\mathcal{B}_i=\text{the set of all person boxes in image }I_i.
\end{equation}

To solve the retrieval problem, the model learns a visual encoder and a text encoder that project visual inputs (image–box pairs) and textual inputs (queries) into a unified $C$-dimensional embedding space:
\begin{equation}
f_v:\ \mathcal{I}\times\mathbb{R}^4\rightarrow\mathbb{R}^C,\qquad
f_t:\ \mathcal{T}\rightarrow\mathbb{R}^C.
\end{equation}

We further define a similarity scoring function to measure cross-modal matching in the shared embedding space:
\begin{equation}
\mathcal{S}:\ \mathbb{R}^C\times\mathbb{R}^C\rightarrow\mathbb{R}.
\end{equation}

Therefore, the core objective of TBPS is that for a positive (matched) pair and a negative (mismatched) pair, the similarity satisfies the following constraint. Let
$b_p=(I_p,\mathbf{bb}_p)\in\mathcal{B}_{gt}(q)$ and
$b_n=(I_n,\mathbf{bb}_n)\notin\mathcal{B}_{gt}(q)$, then
\begin{equation}
\mathcal{S}\!\big(f_t(q),\,f_v(I_p,\mathbf{bb}_p)\big)\;>\;
\mathcal{S}\!\big(f_t(q),\,f_v(I_n,\mathbf{bb}_n)\big).
\end{equation}

As shown in related baselines, the learning process is typically optimized with multi-task losses such as $\mathcal{L}_{\mathrm{cmpm}}$ and $\mathcal{L}_{\mathrm{oim}}$.

\section{Method}

In this section, we first briefly introduce the baseline person search network upon which the proposed \textbf{ICON} is built. Subsequently, we elaborate on the three core components constituting the Neuro-Symbolic Counterfactual Intervention (NeSy-P) framework for Text-Based Person Search (TBPS). These components are Rule-Guided Spatial Intervention, Counterfactual Context Disentanglement, and Saliency-Driven Semantic Regularization. Finally, we present the detailed procedures for the training and inference phases of the framework.

    \begin{figure}[htbp] %H为当前位置，!htb为忽略美学标准，htbp为浮动图形
		\centering %图片居中
		\includegraphics[width=1.0\textwidth]{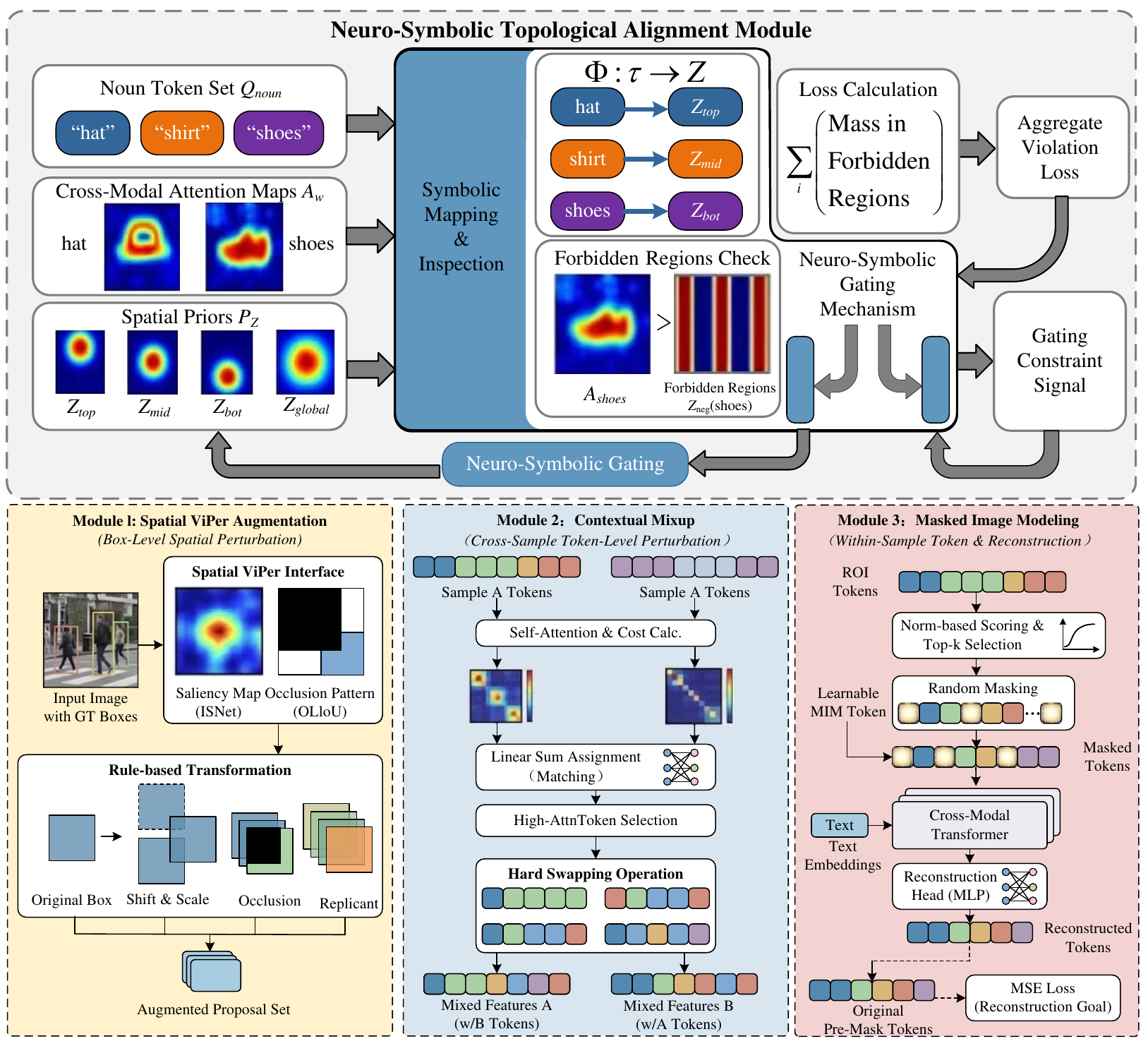} 
		\caption{This is overview} %最终文档中希望显示的图片标题
		\label{Fig.overview} %用于文内引用的标签
	\end{figure}%结束环境

\subsection{Rule-Guided Spatial Intervention}
\label{sec:method_spatial}

Standard TBPS paradigms, such as OIM \citep{xiao2017oim} and SeqNet \citep{li2021seqnet}, typically rely on the assumption that region proposals provided during training are precisely aligned with the target. Consequently, these methods primarily focus on feature extraction within fixed boundaries. However, this approach overlooks unavoidable localization errors in open-world inference, such as misaligned bounding boxes or partial cropping. Although general data augmentations like Random Erasing \citep{zhong2020re} are occasionally employed to enhance robustness, such \textit{Blind Perturbations} are inherently suboptimal. This is because they generate samples indiscriminately, resulting in samples that are either too simple to provide meaningful gradient signals or geometrically invalid, effectively acting as noise rather than regularization. To bridge this gap and sever the \textit{Location Shortcut}, we propose \textbf{Rule-Guided Spatial Intervention}. This is a neuro-symbolic module designed to shift the paradigm from stochastic augmentation to adversarial selection.

To implement this shift, we formally formulate spatial intervention as a constrained optimization problem. First, given a ground-truth bounding box $B_{gt}$, we explicitly construct a candidate pool $\mathcal{B}$ via discrete shifting and scaling operations. Then, instead of randomly sampling from this pool, we employ a neuro-symbolic scoring function $\mathcal{J}$ to identify the optimal intervention sample $B^*$. This process aims to balance adversarial difficulty with geometric validity, with the objective function expressed as $B^* = \operatorname*{argmax}_{B_k \in \mathcal{B}} \mathcal{J}(B_k)$. The specific scoring calculation depends on the following three synergistically designed principles.

This scoring mechanism is driven by three principles derived from symbolic rules, which enforce constraints on semantics, geometry, and stability, respectively. First, to sever the model's dependency on dominant features, the intervention must maximize the \textbf{Semantic Information Loss} ($\mathcal{S}_{adv}$). We quantify this by calculating the drop in attention mass within the bounding box using the model's current attention map $\mathbf{A}$, as a larger drop implies that the candidate successfully crops out high-confidence regions. Second, since maximizing visual loss alone may cause the sample to degenerate into background noise, we simultaneously enforce \textbf{Geometric Realism} ($\mathcal{S}_{geo}$). We achieve this by imposing fuzzy logic constraints on the Intersection-over-Union (IoU) and the Visibility approximated using a triangular window function. Third, to prevent model instability during the early training phase, we introduce a \textbf{Curriculum Stability} factor ($\mathcal{S}_{stab}$) derived from the current epoch $t$ and object scale $(w, h)$:
\begin{equation}
    \mathcal{S}_{stab}(t, B_{gt}) = \min\left(1, \frac{t}{T_{warm}}\right) \cdot \mathbb{I}(\min(w, h) > \tau)
\end{equation}
where $\mathbb{I}(\cdot)$ denotes the indicator function, and $\tau$ represents the minimum scale threshold. This term dynamically suppresses aggressive interventions when the object is too small or during the warm-up period $T_{warm}$. Consequently, by training on these carefully selected samples, the model is forced to internalize \textit{Geometric Invariance}, thereby maintaining consistent recognition capabilities even in the presence of significant localization noise.

\subsection{Counterfactual Context Disentanglement}

In addition to localization noise, existing alignment frameworks \citep{chen2020nae,su2024maca} typically treat the bounding box content as a holistic signal. This approach inadvertently encourages the model to exploit background correlations for retrieval. For instance, NAE \citep{chen2020nae} attempts to suppress background noise via norm-aware embedding. However, this remains a passive filtering approach that fails to distinguish between informative context and spurious statistical biases (e.g., erroneously associating backpacks with pavement). Without active intervention, the model inevitably falls into the \textit{Context Shortcut}, failing to recognize targets when the environmental context shifts. To address this issue, we propose \textbf{Counterfactual Context Disentanglement}. This is a causal intervention mechanism that forces the model to decouple person-specific semantics from background noise by investigating whether the model recognizes the target even if the surrounding environment is altered.

Specifically, we implement this via a neuro-symbolic attention-guided transplantation strategy, letting $\mathbf{F}$ denote the flattened visual feature sequence. First, we perform \textbf{Dynamic Saliency Filtering} to precisely separate the person from their context. Utilizing the model's self-attention map $\mathbf{A}$ as a semantic guide, we dynamically rank tokens based on their activation levels. In this process, high-weight tokens are inferred as the semantic foreground, while low-weight tokens are regarded as context. This step generates a binary disentanglement mask $\mathbf{M}$ for subsequent intervention, thereby achieving the preliminary decoupling of foreground and background.

Following separation, we employ \textbf{Structure-Aware Context Transplantation} to synthesize counterfactual samples. To ensure that the intervention remains statistically valid within the feature manifold, we avoid random pairing when selecting donor backgrounds and instead adopt an \textbf{Optimal Transport} strategy. This is because random pairing may lead to geometric structural mismatches, whereas Optimal Transport seeks an optimal permutation $\pi^*$ within the mini-batch to minimize the structural cost between the target sample $i$ and the donor sample $\pi(i)$:
\begin{equation}
    \pi^* = \operatorname*{argmin}_{\pi \in \Pi_B} \sum_{i=1}^B \| \mathbf{A}_i - \mathbf{A}_{\pi(i)} \|_2^2
\end{equation}
Consequently, this assignment mechanism effectively ensures structural compatibility between the foreground and the transplanted background, laying the foundation for high-quality sample synthesis.

Based on the aforementioned optimal matching results, we construct the counterfactual feature map $\mathbf{F}_{cf}$ by transplanting the donor context onto the target foreground:
\begin{equation}
    \mathbf{F}_{cf} = \mathbf{M} \odot \mathbf{F}_{target} + (1 - \mathbf{M}) \odot \mathbf{F}_{\pi^*(target)}
\end{equation}
where $\odot$ denotes element-wise multiplication. Finally, by forcing the model to predict consistent identities for both the original $\mathbf{F}_{target}$ and the counterfactual $\mathbf{F}_{cf}$, we explicitly break the statistical dependency between the person and the environment. This process ensures that recognition is anchored solely on person-centric attributes, thereby significantly enhancing the model's robustness in complex environments.

\subsection{Saliency-Driven Semantic Regularization}
\label{sec:method_proto}

While the aforementioned modules effectively address geometric and environmental dependencies, TBPS models still suffer profoundly from \textit{Visual Laziness}, also known as the \textit{Saliency Shortcut}. Driven by the attention mechanism, models tend to fixate on the most dominant features (e.g., a bright red shirt) while suppressing secondary but critical attributes, such as boots or logos. Although recent works like IRRA \citep{jiang2023irra} employ Masked Image Modeling to encourage global context learning, they typically allow reconstruction within the visual modality itself. This creates a leakage channel that allows the model to extrapolate missing patches merely from the statistical patterns of neighboring pixels without truly understanding the semantic content. Consequently, this approach effectively negates the need for deep linguistic reasoning, causing the model to bypass genuine cross-modal semantic interaction. To promote \textit{Holistic Completeness}, we propose \textbf{Saliency-Driven Semantic Regularization}. This is an active intervention strategy that compels the model to reconstruct complete semantics strictly based on cross-modal evidence.

Specifically, we first introduce \textbf{Adversarial Saliency Occlusion}, which is distinct from standard methods employing random masking. We compute token-wise saliency scores $\mathcal{S}_{sal} \in \mathbb{R}^L$ based on feature norms and self-attention weights, aiming to identify the regions the model is currently most confident about. Subsequently, rather than masking trivial background regions, we systematically mask the top-$k$ most salient tokens. This operation generates a corrupted feature view $\tilde{\mathbf{F}}_{vis}$ devoid of simple shortcuts. This adversarial removal forces the model to activate dormant neurons responsible for secondary features, thereby ensuring that the feature representation is not dominated by a single salient attribute.

Crucially, to verify that the model truly understands semantics rather than merely memorizing pixel patterns, we formulate the recovery process as \textbf{Cross-Modal Counterfactual Reasoning}. We employ a lightweight decoder that attempts to reconstruct the original visual features $\mathbf{F}_{orig}$ from the corrupted view $\tilde{\mathbf{F}}_{vis}$, imposing a strict constraint: the reconstruction must be conditioned on the textual description $\mathbf{T}$. The regularization objective is formulated as:
\begin{equation}
    \mathcal{L}_{reg} = \| \mathcal{D}(\tilde{\mathbf{F}}_{vis}, \mathbf{T}) - \mathbf{F}_{orig} \|^2
\end{equation}
where $\mathcal{D}$ denotes the cross-modal decoder. This objective establishes a \textit{Causal Bottleneck}. Since dominant visual cues are occluded, the model is compelled to reason through the text (e.g., utilizing the word ``shirt'' to infer missing color and texture) to minimize the reconstruction error. Thus, this successfully severs the Saliency Shortcut, ensuring that the model's representation is both holistic and semantically aligned with the language description.

\subsection{Uncertainty-Aware Prototype Alignment}
\label{sec:method_proto}

The intervention mechanisms in Sections~\ref{sec:method_spatial}--\ref{sec:method_proto} reshape the training distribution by perturbing spatial layouts, contextual background, and salient semantic regions. However, the alignment objective that ties visual and textual modalities together is still typically optimized under an uniform sample assumption: all positive pairs are treated as equally reliable. In realistic TBPS scenarios, this assumption is fragile. Noisy labels, hard positives, and heavily perturbed views often deviate significantly from the underlying identity manifold. When such samples are forced to participate in distribution matching with the same strength as clean, prototype-consistent samples, the embedding space can be pulled towards spurious directions and the benefits of causal interventions are partially offset.

To mitigate this issue, we introduce an Uncertainty-Aware Prototype Alignment strategy that estimates sample reliability from an identity-centric geometric perspective and uses it to re-weight cross-modal alignment. Let $f_i \in \mathbb{R}^d$ denote the $L_2$-normalized feature of sample $i$ (either an RoI feature or a text embedding), and let $\mathrm{pid}(i)$ be its person identity. For each identity $c$, we maintain a mini-batch prototype
\begin{equation}
\mu_c = \frac{1}{|\mathcal{S}_c|} \sum_{i \in \mathcal{S}_c} f_i,
\quad
\mu_c \leftarrow \frac{\mu_c}{\|\mu_c\|_2},
\label{eq:proto_def}
\end{equation}
where $\mathcal{S}_c$ denotes the set of samples with identity $c$. We then measure the deviation of sample $i$ from its prototype (e.g., via cosine distance), normalize deviations within the batch, and transform them into confidence scores $u_i$ such that samples closer to $\mu_{\mathrm{pid}(i)}$ receive higher scores. Finally, we obtain normalized alignment weights
\begin{equation}
w_i \propto (u_i + \epsilon)^{\gamma_u},
\qquad
\sum_i w_i = 1,
\label{eq:uncertainty_weight}
\end{equation}
where $\gamma_u \in [0,1]$ controls the strength of uncertainty-aware re-weighting: $\gamma_u = 0$ recovers uniform weighting, while larger values progressively emphasize prototype-consistent samples.

In practice, these weights are computed independently on the image and text branches and injected into two key components of ICON. For masked image modeling (Section~\ref{sec:method_mim}), per-sample reconstruction losses are aggregated using image-side weights $w_i^{\text{img}}$. This biases the cross-modal reconstruction objective towards views that are geometrically coherent within each identity, while outlier or severely perturbed samples are softly down-weighted instead of being discarded by hard heuristics. For Similarity Distribution Matching, we analogously obtain image-side and text-side weights $(w_i^{\text{img}}, w_j^{\text{txt}})$ and use them to modulate the bidirectional distribution alignment between RoI embeddings and text embeddings. Identities whose visual and textual prototypes are well-formed contribute more strongly to shaping the joint similarity distribution, whereas noisy pairs have limited influence.

The proposed uncertainty-aware prototype alignment operates as a lightweight, plug-in mechanism on top of our spatial, contextual, and semantic interventions. It requires no architectural changes to the backbone and incurs negligible computational overhead, since prototype computation and weight normalization are confined to each mini-batch. Empirically, this strategy stabilizes training under strong interventions, suppresses the impact of unreliable positives on distribution matching, and yields more coherent identity clusters in the shared embedding space, thereby tightening the link between causal interventions and their effect on cross-modal alignment.

\section{Experiments}
\subsection{Implementation Details}

The detection hyperparameters strictly follow ViPer’s public configuration, and the ROI Align output spatial size is \(16\times 8\). The cross-modal Transformer uses 4 multi-head cross-attention layers with 16 heads. During training, the batch size is 8, the optimizer is Adam, the base learning rate is \(1\times 10^{-5}\), and a linear warm-up is applied over the first \(10\%\) of iterations; the total number of training steps is \(20{,}000\), and the learning rate is decayed by a factor of 10 at the \(12/20\) point of training. Consistent with ViPer, the circular queue sizes in the OIM loss are 5000 (CUHK-SYSU-TBPS) and 500 (PRW-TBPS), with temperature \(\sigma=1/30\) and momentum coefficient \(0.5\); for Spatial ViPer, \(\lambda_1,\lambda_2,n\) are set to \(0.4,0.2,3\), respectively; the minimum/maximum counts of removable and exchangeable tokens \(m_r^0/m_r^1\) and \(m_e^0/m_e^1\) are \(1/4\) and \(4/8\); for Fine-grained ViPer, the masking ratio is \(r=0.5\). At test time, images are resized to a fixed resolution of \(1500\times 900\) pixels. The implementation matches the run script: \texttt{NUM\_WORKERS} is set to 0, the random seed is 42, and weights are saved every 1000 steps. All experiments are conducted on two NVIDIA GeForce RTX 5090D GPUs, using PyTorch~2.8.0+cu129 and Detectron2~0.6.

\subsection{Datasets}

\textbf{PRW-TBPS} is an extension of the PRW dataset tailored for text-driven person search. In the training and query splits, each annotated person bounding box is paired with one or more natural-language sentences. The training set contains 5{,}704 scene images captured by multiple on-campus cameras, covering 483 identities and 14{,}897 person boxes; the query split provides 2{,}057 query boxes from 450 identities with independently annotated text. For evaluation, 6{,}112 scene images that do not overlap with the training set are used as the retrieval gallery.

\textbf{CUHK-SYSU-TBPS} builds on CUHK-SYSU by incorporating language annotations from CUHK-PEDES to form a text-guided person search benchmark. The training split comprises 11{,}206 scene images, 15{,}080 person boxes, and 5{,}532 persons, while the gallery split contains 6{,}978 scene images with 2{,}900 persons. Each person box in the training and query sets is equipped with two independent sentence descriptions. The test protocol follows CUHK-SYSU, varying the gallery size from 50 to 4000 to evaluate performance under different retrieval scales.

\textbf{Evaluation Metrics} Following the protocol in~\citep{sdpg2023_mmw}, we employ mean Average Precision (mAP) and Cumulative Matching Characteristics (CMC Top-$K$) for evaluation. A predicted bounding box is considered a true positive only if it matches the query identity with an $\text{IoU} \ge 0.5$, thereby reflecting the joint performance of detection and retrieval. In addition to Top-1, we report Top-5 and Top-10 to assess retrieval robustness against slight ranking errors, simulating real-world scenarios where operators verify a limited list of candidates.

\subsection{Baselines}

This section collects representative baselines spanning classic joint detection–ReID frameworks, norm/angle decoupling, anchor-free and transformer-based one-step person search, as well as text-driven proposal generation, semantic-aware matching, memory-aided coarse-to-fine alignment, CLIP-based strong baselines, and visual-perturbation training, thereby covering key TBPS challenges including detection–representation conflict, semantic alignment, occlusion, and the train–test distribution gap.

\noindent\textbf{\(\bullet\)} \textbf{OIM}~\citep{xiao2017oim}: End-to-end person search jointly learns detection and identification with the Online Instance Matching loss on CUHK-SYSU, establishing a widely adopted baseline.

\noindent\textbf{\(\bullet\)} \textbf{NAE}~\citep{chen2020nae}: Norm-Aware Embedding decouples feature norm and angle (identity discrimination) to mitigate the objective conflict between detection and ReID.

\noindent\textbf{\(\bullet\)} \textbf{AlignPS}~\citep{yan2021alignps}: An efficient one-stage, anchor-free FCOS-based person search baseline that addresses feature misalignment between detection regions and ReID embeddings.

\noindent\textbf{\(\bullet\)} \textbf{PSTR}~\citep{cao2022pstr}: A transformer-based, end-to-end one-step person search framework that formulates detection and ReID as sequence prediction.

\noindent\textbf{\(\bullet\)} \textbf{SDPG}~\citep{sdpg2023_mmw}: Semantic-Driven Proposal Generation introduces text-guided proposals for TBPS, enabling full-image search using natural-language queries.

\noindent\textbf{\(\bullet\)} \textbf{BSL}~\citep{bsl2024_arxiv}: Semantic-Aware Matching performs full-image text-based person search with semantic cues to better align language with visual regions.

\noindent\textbf{\(\bullet\)} \textbf{MACA}~\citep{su2024maca}: Memory-aided Coarse-to-fine Alignment strengthens text–vision alignment for TBPS via a memory mechanism and progressive matching.

\noindent\textbf{\(\bullet\)} \textbf{CLIP-TBPS}~\citep{cao2024tbpsclip}: An empirical study showing that carefully tuned CLIP constitutes a strong TBPS baseline, highlighting the value of vision–language pretraining.

\noindent\textbf{\(\bullet\)} \textbf{ViPer}~\citep{zhang2025viper}: Visual Perturbation training that simulates spatial misalignment, occlusion, and background clutter during training to reduce the train–inference discrepancy in TBPS.

\subsection{Comparison Experiments}

As summarized in Table~\ref{tab:Comparison Experiments}, our model consistently outperforms previous TBPS baselines on both PRW-TBPS and CUHK-SYSU-TBPS. Early pipelines that couple OIM/NAE/BSL with BiLSTM or BERT backbones already show a clear performance progression, yet even the strongest BSL+BERT variant is surpassed by a large margin: our model improves mAP by more than $11.6$ points on PRW-TBPS and $14.9$ points on CUHK-SYSU-TBPS, together with substantial gains in top-1 accuracy. This indicates that simply strengthening the text encoder is insufficient to bridge the gap caused by semantic blind spots and shortcut biases; explicitly modeling cross-modal causality brings an additional level of benefit on top of stronger backbones.

When compared with recent CLIP-based TBPS frameworks such as SDPG, MACA, and ViPer, our method achieves the best results across all metrics and both datasets. In particular, relative to the strongest baseline ViPer, our approach yields an increase of $0.93$ and $3.46$ mAP, as well as $2.50$ and $3.66$ top-1 accuracy on PRW-TBPS and CUHK-SYSU-TBPS, respectively. The standard deviations remain small, demonstrating that these gains are stable rather than random fluctuations. Overall, the comparison experiments verify that the proposed intervention-driven training strategy can further enhance cross-modal discriminability beyond existing architectural advances.

\begin{table}[t]
\centering
\caption{Comparison Experiments. The best results are highlighted in bold.}
\label{tab:Comparison Experiments}
\renewcommand{\arraystretch}{1.2}

\setlength{\tabcolsep}{1.2pt}

\resizebox{\linewidth}{!}{
\begin{tabular}{l cccc cccc} 
\toprule
 & \multicolumn{4}{c}{PRW-TBPS} & \multicolumn{4}{c}{CUHK-SYSU-TBPS} \\
\cmidrule(lr){2-5} \cmidrule(lr){6-9}
Method & mAP & top-1 & top-5 & top-10 & mAP & top-1 & top-5 & top-10 \\
\midrule
OIM+BiLSTM & $4.50 \pm 0.38$ & $6.54 \pm 0.41$ & $16.13 \pm 0.42$ & $22.60 \pm 0.43$ & $23.36 \pm 0.26$ & $17.24 \pm 0.20$ & $38.01 \pm 0.13$ & $49.98 \pm 0.17$\\
NAE+BiLSTM & $5.26 \pm 0.28$ & $7.40 \pm 0.40$ & $16.91 \pm 0.46$ & $24.52 \pm 0.75$ & $23.18 \pm 0.32$ & $16.95 \pm 0.33$ & $38.87 \pm 0.25$ & $50.20 \pm 0.28$\\
BSL+BiLSTM & $3.54 \pm 0.24$ & $6.31 \pm 0.52$ & $15.71 \pm 0.35$ & $22.70 \pm 0.39$ & $27.40 \pm 0.21$ & $20.64 \pm 0.22$ & $41.87 \pm 0.31$ & $51.38 \pm 0.21$\\
\midrule
OIM+BERT & $8.36 \pm 0.47$ & $14.64 \pm 0.57$ & $31.18 \pm 0.17$ & $40.23 \pm 0.56$ & $43.91 \pm 0.35$ & $35.97 \pm 0.52$ & $61.03 \pm 0.14$ & $71.55 \pm 0.26$\\
NAE+BERT & $9.31 \pm 0.31$ & $14.26 \pm 0.21$ & $31.02 \pm 0.36$ & $39.42 \pm 0.63$ & $46.52 \pm 0.34$ & $38.64 \pm 0.20$ & $65.86 \pm 0.25$ & $75.24 \pm 0.42$\\
BSL+BERT & $10.85 \pm 0.48$ & $16.63 \pm 0.32$ & $35.29 \pm 0.41$ & $46.15 \pm 0.71$ & $49.31 \pm 0.39$ & $40.27 \pm 0.38$ & $66.50 \pm 0.15$ & $75.51 \pm 0.32$\\
SDPG & $11.79 \pm 0.54$ & $22.02 \pm 0.46$ & $43.22 \pm 0.26$ & $53.72 \pm 0.53$ & $49.59 \pm 0.48$ & $48.51 \pm 0.20$ & $75.73 \pm 0.14$ & $82.59 \pm 0.25$\\
MACA & $18.38 \pm 0.22$ & $32.86 \pm 0.49$ & $52.90 \pm 0.36$ & $61.21 \pm 0.66$ & $57.70 \pm 0.54$ & $52.34 \pm 0.32$ & $76.49 \pm 0.15$ & $83.43 \pm 0.21$\\
ViPer & $21.56 \pm 0.42$ & $34.24 \pm 0.39$ & $53.46 \pm 0.31$ & $61.98 \pm 0.63$ & $60.71 \pm 0.36$ & $54.82 \pm 0.37$ & $78.25 \pm 0.21$ & $84.97 \pm 0.34$\\
\midrule
\textbf{ICON} & $\mathbf{22.49} \pm 0.31$ & $\mathbf{36.74} \pm 0.24$ & $\mathbf{53.98} \pm 0.19$ & $\mathbf{62.19} \pm 0.41$ & $\mathbf{64.17} \pm 0.34$ & $\mathbf{58.48} \pm 0.24$ & $\mathbf{80.49} \pm 0.15$ & $\mathbf{86.86} \pm 0.31$ \\
\bottomrule
\end{tabular}
}
\end{table}

\subsection{Ablation Study}

\begin{table}[H]
\centering
\caption{Ablation study.}
\label{tab:ablation_ICON}
\resizebox{\linewidth}{!}{
\begin{tabular}{lcccc cccc cccc}
\toprule
 & & & & & \multicolumn{4}{c}{\textbf{PRW-TBPS}} & \multicolumn{4}{c}{\begin{tabular}{@{}c@{}}\textbf{CUHK-SYSU-TBPS}\end{tabular}} \\
\cmidrule(lr){6-9} \cmidrule(lr){10-13}
\textbf{Variant} & \textbf{A} & \textbf{B} & \textbf{C} & \textbf{D} & \textbf{mAP} & \textbf{top-1} & \textbf{top-5} & \textbf{top-10} & \textbf{mAP} & \textbf{top-1} & \textbf{top-5} & \textbf{top-10} \\
\midrule
FULL  & \checkmark & \checkmark & \checkmark & \checkmark & 22.49 & 36.74 & 53.98 & 62.19 & 64.17 & 58.48 & 80.49 & 86.86 \\
w/o A & --         & \checkmark & \checkmark & \checkmark & 21.13 & 35.08 & 53.35 & 61.89 & 61.12 & 56.16 & 79.05 & 85.11 \\
w/o B & \checkmark & --         & \checkmark & \checkmark & 20.64 & 34.67 & 53.11 & 61.54 & 60.48 & 56.68 & 78.87 & 84.68 \\
w/o C & \checkmark & \checkmark & --         & \checkmark & 20.82 & 34.81 & 53.76 & 61.67 & 61.08 & 56.11 & 79.17 & 84.91 \\
w/o D & \checkmark & \checkmark & \checkmark & --         & 21.32 & 35.56 & 53.83 & 62.20 & 61.45 & 56.71 & 78.92 & 84.65 \\
\bottomrule
\end{tabular}
}
\end{table}

As shown in Table~\ref{tab:ablation_ICON}, we ablate the four components A--D of the proposed NeSy-P framework on both PRW-TBPS and CUHK-SYSU-TBPS. Starting from the \emph{FULL} configuration with all modules enabled, removing the rule-guided spatial intervention (A, Sec.~4.1) already leads to a noticeable performance drop: mAP decreases from 22.49 to 21.13 and top-1 from 36.74 to 35.08 on PRW-TBPS, while on CUHK-SYSU-TBPS mAP drops from 64.17 to 61.12 and top-1 from 58.48 to 56.16. Disabling the cross-sample counterfactual context disentanglement (B, Sec.~4.2) further deteriorates the retrieval quality, with mAP reduced to 20.64 and top-1 to 34.67 on PRW-TBPS, and a similar degradation from 64.17 to 60.48 mAP and from 58.48 to 56.68 top-1 on CUHK-SYSU-TBPS. A comparable decline is observed when turning off the saliency-driven semantic regularization (C, Sec.~4.3): the model falls back to 20.82 mAP / 34.81 top-1 on PRW-TBPS and 61.08 mAP / 56.11 top-1 on CUHK-SYSU-TBPS. These consistent drops across two benchmarks confirm that all three perturbation branches contribute positively and complementarily to the overall robustness of NeSy-P.

These quantitative trends are consistent with the design intent of modules A--D. Module~A explicitly optimizes over an adversarial yet geometrically valid pool of perturbed boxes, forcing the network to preserve identity discrimination under strong localization noise rather than overfitting to a single ground-truth box; ablating A reintroduces the ``location shortcut'', making the model fragile to detector errors and thus lowering mAP. Module~B performs attention-guided foreground/background disentanglement combined with structure-aware context transplantation, so that a person can still be recognized when the surrounding scene changes; once B is removed, the network again exploits spurious co-occurrences between identity and background, which particularly hurts performance on CUHK-SYSU-TBPS where scene bias is strong. Module~C uses saliency-driven adversarial masking and cross-modal reconstruction to suppress ``saliency laziness'': the model is required to recover masked visual tokens purely from textual evidence, encouraging holistic, attribute-level grounding; removing C causes the representation to be dominated by a few over-salient cues and weakens cross-modal semantic alignment, so both mAP and top-$k$ accuracy consistently drop. Finally, module~D introduces an uncertainty-aware alignment mechanism that reweights the SDM and MIM objectives at the instance level, down-weighting noisy or ambiguous pairs while emphasizing confident, class-consistent regions; disabling D makes the optimization more susceptible to outliers and annotation noise, slightly reducing the overall gains even though the base contrastive objective remains. Taken together, A--D form a coherent set of causal interventions and instance-level reweighting strategies that systematically close the train--test gap under misalignment, occlusion, and background shift.

\subsection{Hyperparameter Experiment}

\begin{figure}[H] %H为当前位置，!htb为忽略美学标准，htbp为浮动图形
		\centering %图片居中
		\includegraphics[width=1.0\textwidth]{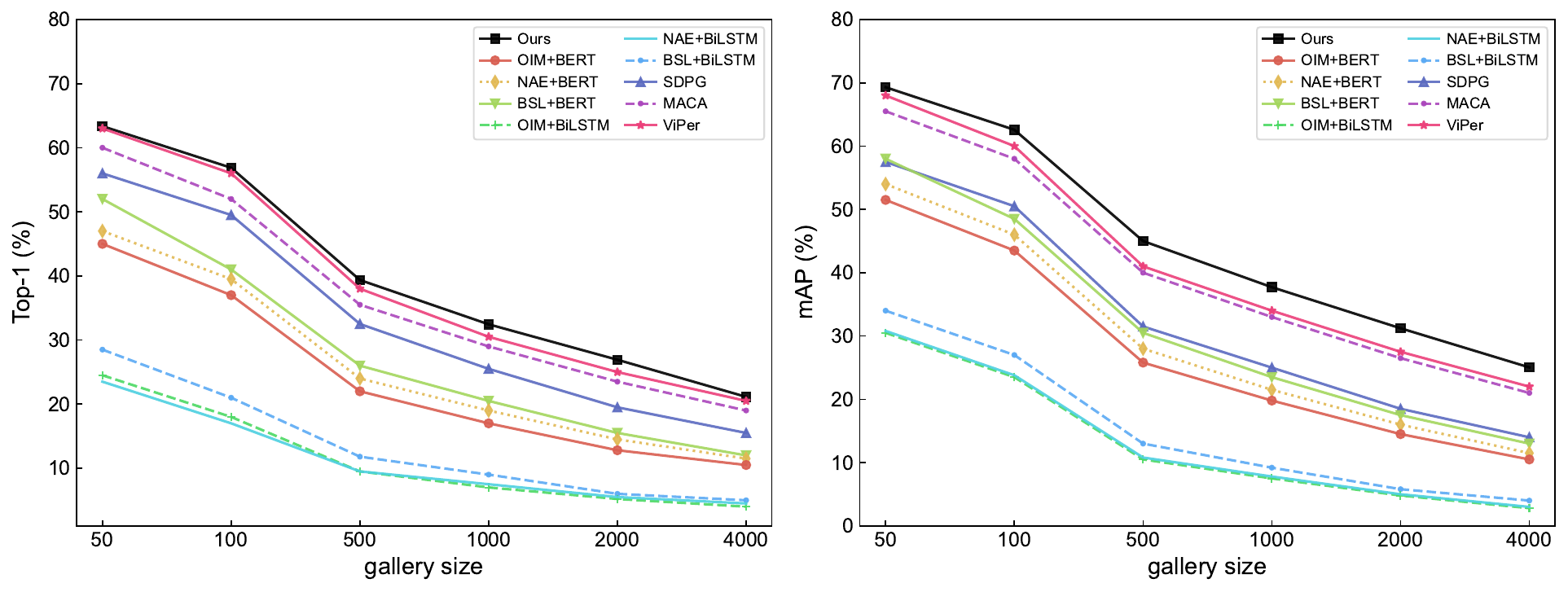} 
		\caption{Person search performances on CUHK-SYSU-TBPS under various gallery sizes.} %最终文档中希望显示的图片标题
		\label{Fig.gallery_size_comparison} %用于文内引用的标签
	\end{figure}

As illustrated in Figure~\ref{Fig.gallery_size_comparison}, we investigate the scalability of our model on the CUHK-SYSU-TBPS benchmark by varying the gallery size from $50$ to $4,000$. It is observed that while performance metrics (Top-1 and mAP) naturally decline for all methods as the search space expands due to the increased presence of distractors, our proposed framework consistently outperforms state-of-the-art competitors (e.g., ViPer and MACA) across the entire spectrum. Notably, our method maintains a distinct performance margin even at the largest gallery size ($N=4,000$), demonstrating its superior robustness in handling complex, large-scale retrieval scenarios where distinguishing the target from massive environmental noise is critical.

\begin{figure}[H] %H为当前位置，!htb为忽略美学标准，htbp为浮动图形
		\centering %图片居中
		\includegraphics[width=1.0\textwidth]{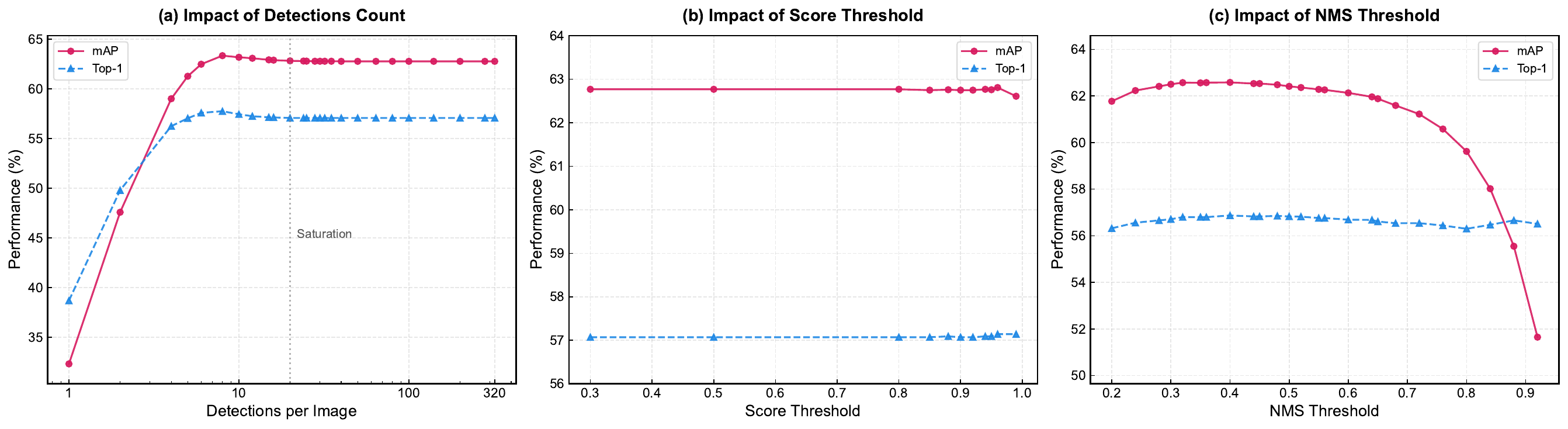} 
		\caption{Impact of inference hyperparameters on model performance. We illustrate the variations in mAP and Top-1 accuracy under different settings of (a) maximum detections per image, (b) confidence score threshold, and (c) NMS IoU threshold. The curves highlight the optimal operating ranges for each parameter.} %最终文档中希望显示的图片标题
		\label{Fig.hyperparameter} %用于文内引用的标签
	\end{figure}

As illustrated in Figure~\ref{Fig.hyperparameter}, we evaluate the model's robustness against variations in three key inference hyperparameters. First, regarding \textbf{detections per image} (Figure~\ref{Fig.hyperparameter}(a)), performance improves rapidly as the number of detections increases from $1$ to $8$ before reaching a saturation plateau. This trajectory evidences the model's high recall efficiency, where optimal target coverage is achieved with minimal candidate boxes (e.g., top-10), thereby significantly reducing computational overhead compared to redundancy-heavy traditional methods.In terms of filtering parameters, the \textbf{confidence score threshold} (Figure~\ref{Fig.hyperparameter}(b)) exhibits remarkable stability across a wide range ($0.3$ to $0.95$), reflecting a substantial decision margin between targets and background clutter. Meanwhile, the \textbf{NMS threshold} (Figure~\ref{Fig.hyperparameter}(c)) achieves peak performance between $0.4$ and $0.5$, though accuracy degrades at excessively high thresholds ($>0.7$) due to insufficient suppression of overlapping boxes. Collectively, these results demonstrate that our framework maintains strong robustness across standard parameter ranges without requiring extensive fine-tuning.

%结束环境

%结束环境

\subsection{Visualization}

\begin{figure}[H]
    \centering
    \includegraphics[width=\linewidth]{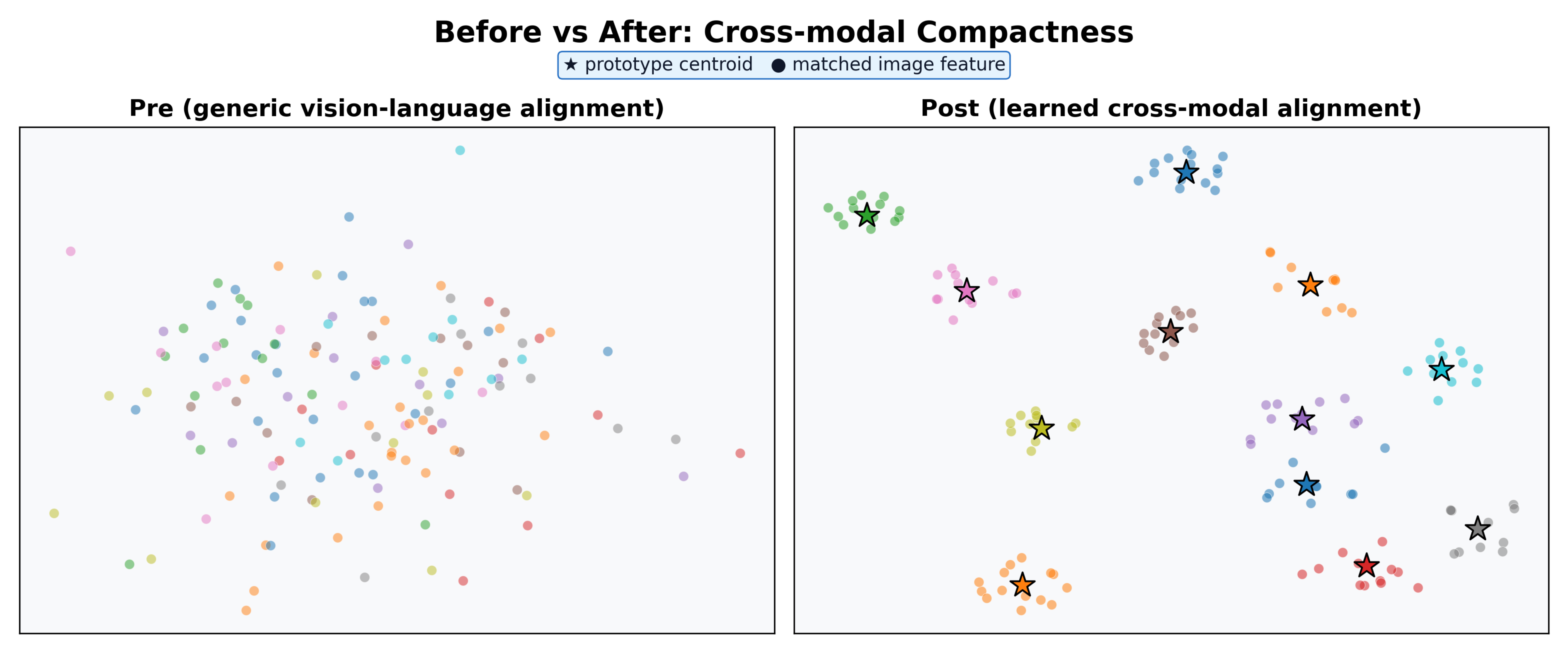}
    \caption{Cross-modal feature distributions before and after training; our method compacts same-ID features around their prototypes and improves cluster separation.}
    \label{fig:vis_compactness}
\end{figure}

To better understand how our interventions reshape the joint vision--language space, we first visualize the cross-modal compactness before and after training. We randomly sample several person IDs and treat their text embeddings as class prototypes. For each ID, we collect all matched image features from the validation set and project both prototypes and image features into a 2D space using UMAP. The left side of Figure~\ref{fig:vis_compactness} shows that, under the generic vision--language alignment, image features from different IDs are heavily entangled and loosely clustered around their prototypes. In contrast, the right side illustrates that our intervention-driven training encourages each prototype to form a well-separated, compact cluster of matched visual features, indicating more faithful and geometry-aware cross-modal alignment.

\begin{figure}[H]
    \centering
    \includegraphics[width=\linewidth]{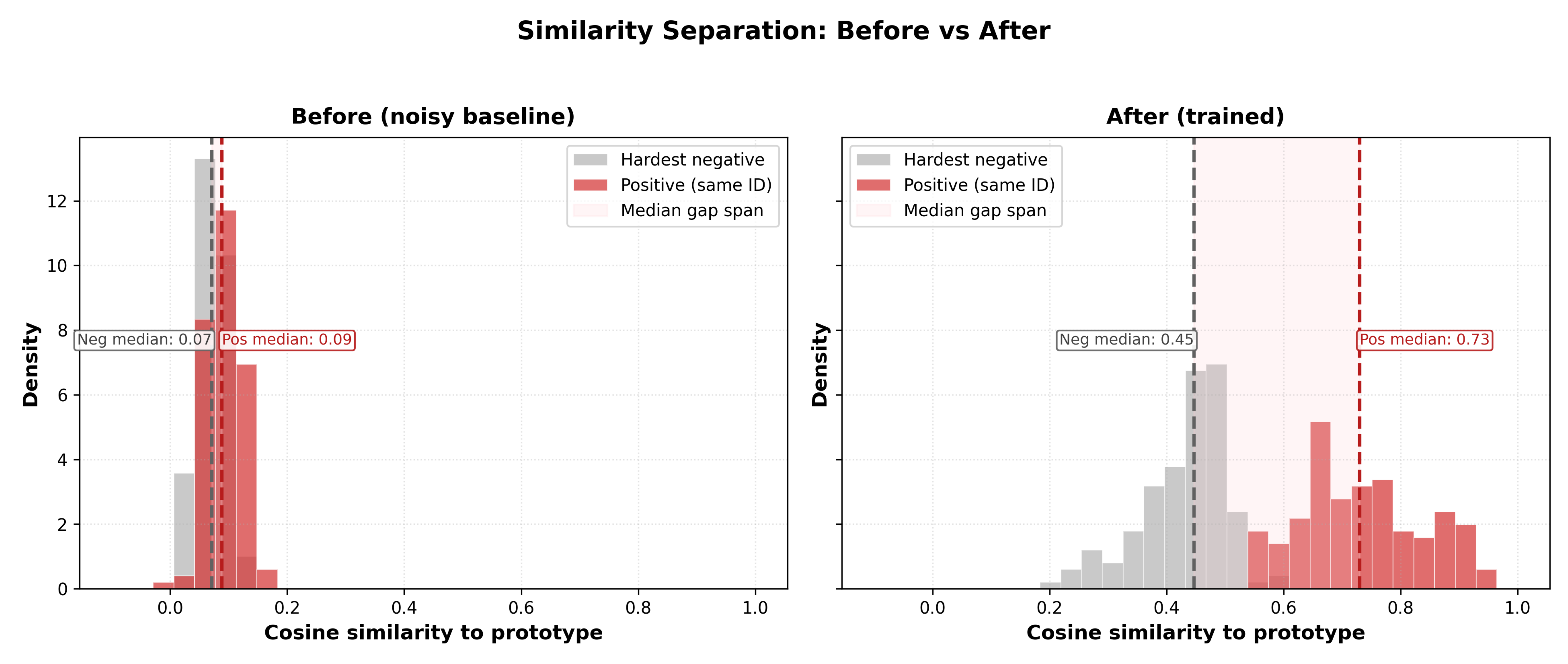}
    \caption{Cosine similarity to text prototypes for positives and hardest negatives before and after training, showing an enlarged median gap and improved discriminability.}
    \label{fig:vis_similarity_gap}
\end{figure}

We further quantify this effect in the similarity space. For the same set of IDs, we compute the cosine similarity between each image feature and its corresponding text prototype, and simultaneously collect the hardest negative for each prototype (i.e., the non-matching image with the highest similarity). Figure~\ref{fig:vis_similarity_gap} compares the similarity distributions before and after training. In the noisy baseline, positive and hardest-negative similarities almost overlap, leading to a very small median gap. After applying our method, the positive distribution shifts significantly to the right while the hardest negatives remain well separated, yielding a much larger median gap. This confirms that our model not only compacts same-ID features around their prototypes, but also enlarges the decision margin against hard negatives, which directly supports the observed gains in retrieval accuracy.

\section{Conclusion}

In this work, we revisited text-based person search from the perspective of semantic causality and proposed ICON, an active-intervention framework that explicitly enforces geometric invariance, environmental independence, and holistic completeness. ICON integrates three complementary perturbation mechanisms and an uncertainty-aware alignment module: rule-guided spatial intervention injects adversarial yet plausible localization noise to suppress location shortcuts; cross-sample contextual perturbation disentangles identity cues from background co-occurrences; saliency-driven semantic regularization prevents over-reliance on a few dominant visual cues by enforcing reconstruction from textual evidence; and instance-wise uncertainty weighting mitigates the impact of noisy or ambiguous supervision during optimization. Extensive experiments on PRW-TBPS and CUHK-SYSU-TBPS demonstrate that ICON consistently improves mAP and top-$k$ retrieval accuracy.


\begin{thebibliography}{00}

\bibitem[Radford et al.(2021)]{radford2021clip}
A. Radford, J.W. Kim, C. Hallacy, A. Ramesh, G. Goh, S. Agarwal, G. Sastry, A. Askell, P. Mishkin, J. Clark, G. Krueger, I. Sutskever,
Learning Transferable Visual Models From Natural Language Supervision.
In: Proceedings of ICML, PMLR 139, pp. 8748--8763 (2021).

\bibitem[He et al.(2016)]{he2016resnet}
K. He, X. Zhang, S. Ren, J. Sun,
Deep Residual Learning for Image Recognition.
In: CVPR, pp. 770--778 (2016).

\bibitem[Ren et al.(2015)]{ren2015faster}
S. Ren, K. He, R. Girshick, J. Sun,
Faster R-CNN: Towards Real-Time Object Detection with Region Proposal Networks.
In: NeurIPS, pp. 91--99 (2015).

\bibitem[Vaswani et al.(2017)]{vaswani2017attention}
A. Vaswani, N. Shazeer, N. Parmar, J. Uszkoreit, L. Jones, A.N. Gomez, L. Kaiser, I. Polosukhin,
Attention Is All You Need.
In: NeurIPS, pp. 5998--6008 (2017).

\bibitem[Dosovitskiy et al.(2021)]{dosovitskiy2021vit}
A. Dosovitskiy, L. Beyer, A. Kolesnikov, D. Weissenborn, X. Zhai, T. Unterthiner, M. Dehghani, M. Minderer, G. Heigold, S. Gelly, J. Uszkoreit, N. Houlsby,
An Image is Worth 16x16 Words: Transformers for Image Recognition at Scale.
In: ICLR (2021).

\bibitem[Jiang and Ye(2023)]{jiang2023irra}
D. Jiang, M. Ye,
Cross-Modal Implicit Relation Reasoning and Aligning for Text-to-Image Person Retrieval.
In: CVPR, pp. 2787--2797 (2023).

\bibitem[Cao et al.(2024)]{cao2024tbpsclip}
M. Cao, Z. Guo, R. Xin, J. Pan, S. Lyu, C. Kang,
An Empirical Study of CLIP for Text-Based Person Search.
In: AAAI, pp. 14405--14413 (2024).

\bibitem[Zhang et al.(2025)]{zhang2025viper}
P. Zhang, X. Yu, X. Bai, J. Zheng,
Visual Perturbation for Text-Based Person Search.
In: AAAI, 39(10), pp. 10058--10066 (2025).


\bibitem[Xiao et al.(2017)]{xiao2017oim}
T. Xiao, S. Li, B. Wang, L. Lin, X. Wang,
Joint Detection and Identification Feature Learning for Person Search.
In: CVPR, pp. 3415--3424 (2017).

\bibitem[Zheng et al.(2017)]{zheng2017prw}
L. Zheng, H. Zhang, S. Sun, M. Chandraker, Y. Yang, Q. Tian,
Person Re-identification in the Wild.
In: CVPR, pp. 1367--1376 (2017).

\bibitem[Chen et al.(2020)]{chen2020nae}
D. Chen, S. Zhang, J. Yang, B. Schiele,
Norm-Aware Embedding for Efficient Person Search.
In: CVPR, pp. 12612--12621 (2020).

\bibitem[Li and Miao(2021)]{li2021seqnet}
Z. Li, D. Miao,
Sequential End-to-end Network for Efficient Person Search.
In: AAAI, 35(3), pp. 2011--2019 (2021).

\bibitem[Bai et al.(2023)]{bai2023noparallel}
Y. Bai, D. Liu, X. Bai, X. Yang, J. Zhou,
Text-based Person Search without Parallel Image-Text Data.
In: ACM MM, pp. 5460--5468 (2023).

\bibitem[Zhong et al.(2020)]{zhong2020re}
Z. Zhong, L. Zheng, G. Kang, S. Li, Y. Yang,
Random Erasing Data Augmentation.
In: AAAI, 34(7), pp. 13001--13008 (2020).

\bibitem[Sun et al.(2018)]{sun2018pcb}
Y. Sun, L. Zheng, Y. Yang, Q. Tian, S. Wang,
Beyond Part Models: Person Retrieval with Refined Part Pooling and a Strong Convolutional Baseline.
In: ECCV, pp. 501--518 (2018).

\bibitem[Miao et al.(2019)]{miao2019pgfa}
J. Miao, Y. Wu, P. Liu, Y. Ding, Y. Yang,
Pose-Guided Feature Alignment for Occluded Person Re-Identification.
In: ICCV, pp. 542--551 (2019).

\bibitem[Tan et al.(2024)]{tan2024spt}
L. Tan, J. Xia, W. Liu, P. Dai, Y. Wu, L. Cao,
Occluded Person Re-Identification via Saliency-Guided Patch Transfer.
In: AAAI, 38, pp. 5070--5078 (2024).

\bibitem[Wu et al.(2024)]{wu2024mgcc}
X. Wu, W. Ma, D. Guo, T. Zhou, S. Zhao, Z. Cai,
Text-based Occluded Person Re-identification via Multi-Granularity Contrastive Consistency Learning.
In: AAAI, 38(6), pp. 6162--6170 (2024).

\bibitem[Han et al.(2021)]{han2021dmrnet}
C. Han, Z. Zheng, K. Su, D. Yu, Z. Yuan, C. Gao, N. Sang, Y. Yang,
Towards Effective Feature Learning for One-Step Person Search.
In: AAAI (2021).

\bibitem[Yan et al.(2021)]{yan2021alignps}
Y. Yan, J. Li, J. Qin, S. Bai, S. Liao, L. Liu, F. Zhu, L. Shao,
Anchor-Free Person Search.
In: CVPR, pp. 7686--7695 (2021).

\bibitem[Cao et al.(2022)]{cao2022pstr}
J. Cao, Y. Pang, R.M. Anwer, H. Cholakkal, J. Xie, M. Shah, F.S. Khan,
PSTR: End-to-End One-Step Person Search With Transformers.
In: CVPR (2022).

\bibitem[Su et al.(2024)]{su2024maca}
L. Su, R. Quan, Z. Qi, R. Fang, Z. Zhang, X. Zhang,
MACA: Memory-aided Coarse-to-fine Alignment for Text-based Person Search.
In: SIGIR (2024).

\bibitem[Zhang et al.(2023)]{sdpg2023_mmw}
C. Zhang, Y. Li, Z. Wang, et al.,
Text-based Person Search in Full Images via Semantic-Driven Proposal Generation.
In: HCMA@MM (Workshop on Human-centric Multimedia Analysis at ACM MM), pp. 5--14 (2023).

\bibitem[Zhang et al.(2024)]{bsl2024_arxiv}
C. Zhang, Y. Li, Z. Wang, et al.,
Text-based Person Search in Full Images via Semantic-Aware Matching.
In: arXiv preprint arXiv:2109.12965 (2024).



\end{thebibliography}
\end{document}